\documentclass{article}

\usepackage[square,numbers]{natbib}
\PassOptionsToPackage{numbers, compress, square}{natbib}
\usepackage[preprint]{neurips_2020}

\usepackage[utf8]{inputenc} % allow utf-8 input
\usepackage[T1]{fontenc}    % use 8-bit T1 fonts
\usepackage{hyperref}       % hyperlinks
\usepackage{url}            % simple URL typesetting
\usepackage{booktabs}       % professional-quality tables
\usepackage{amsfonts}       % blackboard math symbols
\usepackage{nicefrac}       % compact symbols for 1/2, etc.
\usepackage{microtype}
\usepackage{graphicx}      % microtypography
\usepackage{mathtools}
\newcommand{\proofnum}[2]{\vspace{3mm}	\noindent {\sc Proof #1}{\it #2} \vspace{3mm}}
\newtheorem{prop}{Proposition}
\newtheorem{corollary}{Corollary}
\usepackage{amssymb, color, subcaption}                     

\newcommand{\va}{\mathbf{a}}
\newcommand{\vw}{\mathbf{w}}
\newcommand{\vx}{\mathbf{x}}
\newcommand{\tr}{^{\intercal}}
\newcommand{\mW}{\mathbf{W}}
\newcommand{\RR}{\mathbb{R}}
\newcommand{\CI}{\mathcal{I}}
\newcommand{\CA}{\mathcal{A}}
\newcommand{\ve}{\mathbf{e}}
\newcommand{\vz}{\mathbf{z}}
\newcommand{\CO}{\mathcal{O}}

\title{Low-dimensional Interpretable Kernels with \\
  Conic Discriminant Functions for Classification}

\author{%
  Gürhan Ceylan\thanks{Corresponding author.} \\
  Eskişehir Teknik Üniversitesi \\
  Eskişehir, Turkey \\
  \texttt{gurhanceylan@eskisehir.edu.tr} \\
  \And
  Ş. İlker Birbil\\
  Econometric Institute \\
  Erasmus University Rotterdam \\
  3000 DR Rotterdam, The Netherlands\\
  \texttt{birbil@ese.eur.nl} \\
}

\begin{document}

\maketitle

\begin{abstract}

  Kernels are often developed and used as implicit mapping functions
  that show impressive predictive power due to their high-dimensional
  feature space representations. In this study, we gradually construct
  a series of simple feature maps that lead to a collection of
  interpretable low-dimensional kernels. At each step, we keep the
  original features and make sure that the increase in the dimension
  of input data is extremely low, so that the resulting discriminant
  functions remain interpretable and amenable to fast
  training. Despite our persistence on interpretability, we obtain
  high accuracy results even without in-depth hyperparameter
  tuning. Comparison of our results against several well-known kernels
  on benchmark datasets show that the proposed kernels are competitive
  in terms of prediction accuracy, while the training times are
  significantly lower than those obtained with state-of-the-art kernel
  implementations.
\end{abstract}

\section{Introduction}

Whenever the relationship between input and output of a dataset is not
linear, kernels are indispensable tools to model this nonlinearity by
expanding the feature space. This expansion is sometimes carried with
an explicit function called the \textit{feature map}. Capturing the
nonlinear relationship has been observed countless times to have a
profound effect on the estimation accuracy. However, using kernels has
two important caveats from a practitioner's point of view: First, the
dimension of the problem can become quite large after expanding the
feature space. In fact for some of the kernels, even the explicit
feature map is not known. Thus, the interpretation of the trained
models becomes a daunting task. Second, the increase in the problem
dimension causes an increase in the training times. When explicit
feature map is known, then fast linear algorithms can be used
\cite{chang2011libsvm, rahimi2008random}.  However, even in this case,
expanding the feature space to large dimensions can still slow down
the ovearall training process. In this study, we propose several
simple feature maps that lead to a collection of interpretable kernels
with varying degrees of freedom. We make sure that the increase in the
dimension of input data with each proposed feature map is extremely
low, so that the resulting models can be trained quickly, and the
obtained results can easily be interpreted.

Approximating kernels via feature maps is a common approach to achieve
fast training \cite{rahimi2008random, maji2009max,
  kar2012random,vempati2010generalized,
  pham2013fast,hamid2014compact,avron2016quasi,
  si2017memory}. Generally speaking, the success of an approximation
increases as the dimension of the feature map increases. Therefore, a
practitioner needs to conduct a series of experiments to find a
feature map with a desirable performance in terms of accuracy and
training time. When dimension increases, the resulting model becomes
more difficult to interpret. For instance, the approximation method
proposed by \citet{pham2013fast} achieves the same classification
accuracy of a second order polynomial kernel by an almost 10-fold
increase in the dimension. Our work has also ties with piecewise
linear mapping functions introduced by
\citet{huang2013support}. However, like approximation methods, their
results depend on the choice of the dimension. The authors report
numerical results where the increase in the dimension with the
proposed feature maps goes up to 20-fold.

In this study, we center our contributions around two schemes to
construct low-dimensional feature maps. The first scheme increments
the dimension of the input data only by one, whereas the second scheme
doubles dimension of the feature map. These dimensions are much lower
than the existing feature maps in the literature. Our feature maps may
require an anchor point as a hyperparameter. We argue that this
hyperparameter can either be set by the domain experts or by the
training algorithms automatically. Combined with the low
dimensionality, the results obtained with the proposed maps can be
easily interpreted by the practitioners. We also discuss several
conditions under which perfect separability is guaranteed for binary
classification datasets. This observation can also be used to propose
other methods for selecting anchor points. To elaborate on these
points, we reserve a section in the supplementary document about how
to extend the proposed schemes to intermediate dimensions as well as
to multi-class classification.

In a different line of work, \citet{gasimov2006separation} and
\citet{Cevikalp_2017_CVPR} introduce classifiers based on polyhedral
conic functions that are similar to our feature maps. Both papers have
used the polyhedral conic functions within optimization problems but
have not established the relations of these functions to feature maps
and their associated kernels. In a recent follow-up study,
\citet{cevikalp2019polyhedral} have also mentioned the ellipsoidal
conic functions. Here, we consider a general class that includes both
the polyhedral and the ellipsoidal conic functions. We also explicitly
present the resulting feature maps along with the corresponding
kernels. This shows that the related classifiers in the literature
are just kernel methods. Our discussion through explicit feature maps
has far reaching consequences, as associated kernels can be used in
various learning methods.

In the light of this review, we make the following contributions to
the literature:
\begin{itemize}

\item We propose several low-dimensional feature maps, which simply
  concatenates  the original input features with distance-based
  features. The proposed feature maps are easy to interpret and their
  training times are in par with linear kernels.

\item We compare the new kernels against the commonly used kernels on
  a set of binary classification datasets. Our numerical results
  demonstrate that the proposed kernels obtain high accuracy, and on
  several datasets, even outperform all the other kernels. We also
  note that our results with the proposed kernels are obtained without
  any parameter tuning.

\item We demonstrate with our implementation\footnote{(GitHub page) --
    \url{https://github.com/sibirbil/SimpleKernels}} that it is truly
  simple to incorporate the proposed kernels within the existing
  software packages used for kernel-based methods.
  
\end{itemize}

\section{Low-dimensional feature maps and kernels}
\label{sec:proposed}
\begin{figure}
\centering
\includegraphics[scale=0.5]{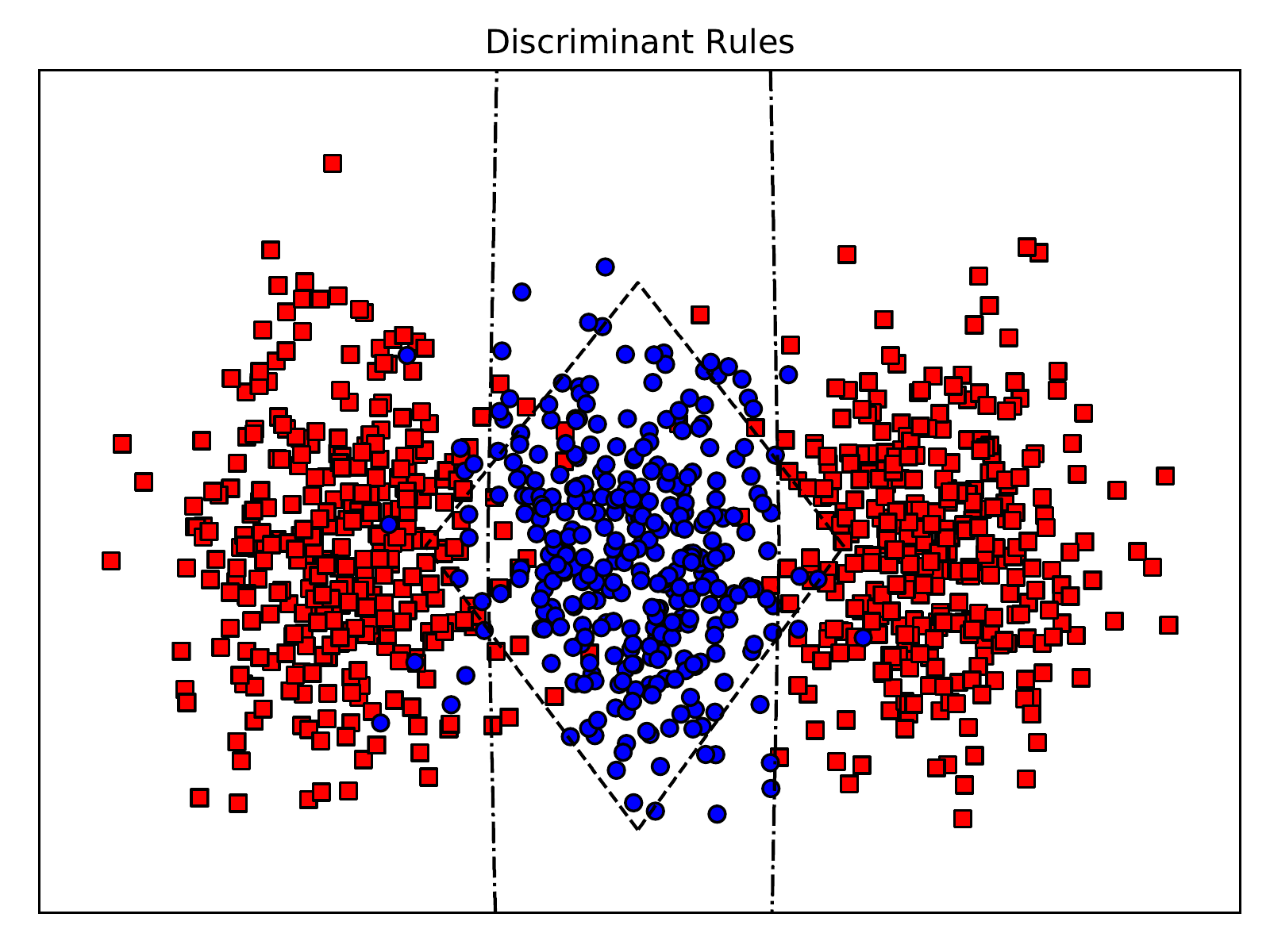}
\caption{Conic discriminant functions obtained with a single anchor
  point for a binary classification dataset. Here, the anchor point is
  taken as the average of all samples.}
\label{fig:diff}
\end{figure}

Consider a practitioner, who trains a classification model on a
dataset to obtain the weights associated with the features of the
$d$-dimensional input vector $\mathbf{x}$. When the relationship is
linear, we obtain the most interpretable model with the discriminant
function\footnote{We omit the bias term added to the discriminant
  function to simplify our exposition.}
\[
  f(\vx) = \vw_{1:d}\tr\vx = w_1x_1 + w_2x_2 + \cdots + w_dx_d,
\]
where $\vw_{1:d}$ stands for the $d$-dimensional weight vector. Our
first model simply asks the practitioner to set an anchor point $\va$
and measure its the distance to the input. That is
\begin{equation}
  \label{eqn:gendisc}
  f_{p,1}(\vx~|~\va) = \vw_{1:d+1}\tr \phi_{p,1}(\vx~|~\va) = \vw_{1:d}\tr\vx +
  w_{d+1}\| \vx - \va\|^p_p,
\end{equation}
where $\phi_{p,1}(\vx~|~\va): \RR^d \mapsto \RR^{d+1}$ with $p > 0$ is
the proposed feature map that adds \textit{only one} dimension, and
$w_{d+1}$ is the weight corresponding to the new feature measuring the
distance between the anchor and the input\footnote{In case $p=\infty$,
  we abuse the notation slightly and work with
  $\ell_\infty$-norm.}. Explicitly, this feature map is given by
\[
  \phi_{p,1}(\vx~|~\va) = (x_1, x_2, \dots, x_d, \|\vx - \va\|^p_p)\tr =
  (\vx\tr, \| \vx - \va\|^p_p)\tr.
\]
Throughout our discussion, we mainly use either $p=1$ or $p=2$, which
are the two most common choices in a wide-range of learning
algorithms. The discriminant function obtained with this feature map
is quite stringent in the sense that only one more weight ($w_{d+1}$)
is added with the anchor point. Thus, one immediate extension could be
using different weights for each feature. In fact, this extension
takes us to our second feature map
$\phi_{p,d}(\vx): \RR^d \mapsto \RR^{2d}$ and the corresponding
discriminant function. For $p=1$, we obtain
\[
  f_{1,d}(\vx~|~\va) = \vw_{1:2d}\tr \phi_{1,d}(\vx~|~\va) = \vw_{1:d}\tr\vx +
  \sum_{\ell=1}^dw_{d+\ell}|x_\ell - a_\ell|.
\]
If we further define a diagonal matrix $\mW$ with elements
$w_{d+1}, w_{d+2} \dots, w_{2d}$, then the discriminant function for
$p=2$ with the second feature map becomes
\[
  f_{2, d}(\vx~|~\va) = \vw_{1:2d}\tr \phi_{2,d}(\vx~|~\va) =
  \vw_{1:d}\tr\vx + (\vx - \va)\tr\mW (\vx - \va).
\]
Contrasting this discriminant function with \eqref{eqn:gendisc} shows
that after doubling the dimension, we still measure a
\textit{distance} to the anchor point. When the elements of $\mW$ are
restricted to be positive, then the last term indeed becomes a
weighted norm. The feature map that doubles the dimension is given by
\[
  \phi_{p,d}(\vx~|~\va) = (x_1, x_2, \dots, x_d, |x_1 - a_1|^p, \dots, |x_d
  - a_d|^p)\tr.
\]
Figure \ref{fig:diff} illustrates two discriminant rules obtained with
feature maps $\phi_{1,1}$ (diamond) and $\phi_{1,d}$ (vertical lines)
for a two-dimensional binary classification problem. The flexibility
of using more dimensions with $\phi_{1,d}$ provides a clear separation
for classification. Note that the regions defined by the rules are the
lower-level sets of different cones pointed at the anchor point. Thus,
we refer to the proposed functions as \textit{conic discriminant
  functions}.  These conic discriminant functions have ties with the
line of work initiated by \citet{gasimov2006separation} and then
extended by \citet{Cevikalp_2017_CVPR}. In these two works, the
discriminant functions are coined as \textit{polyhedral conic
  classifiers}. In Section \ref{sec: pcf}, we elaborate on the
relationship between our current work and the conic classifiers.

\begin{figure}
  \begin{subfigure}{.33\textwidth}
    \centering
    \includegraphics[scale=0.25]{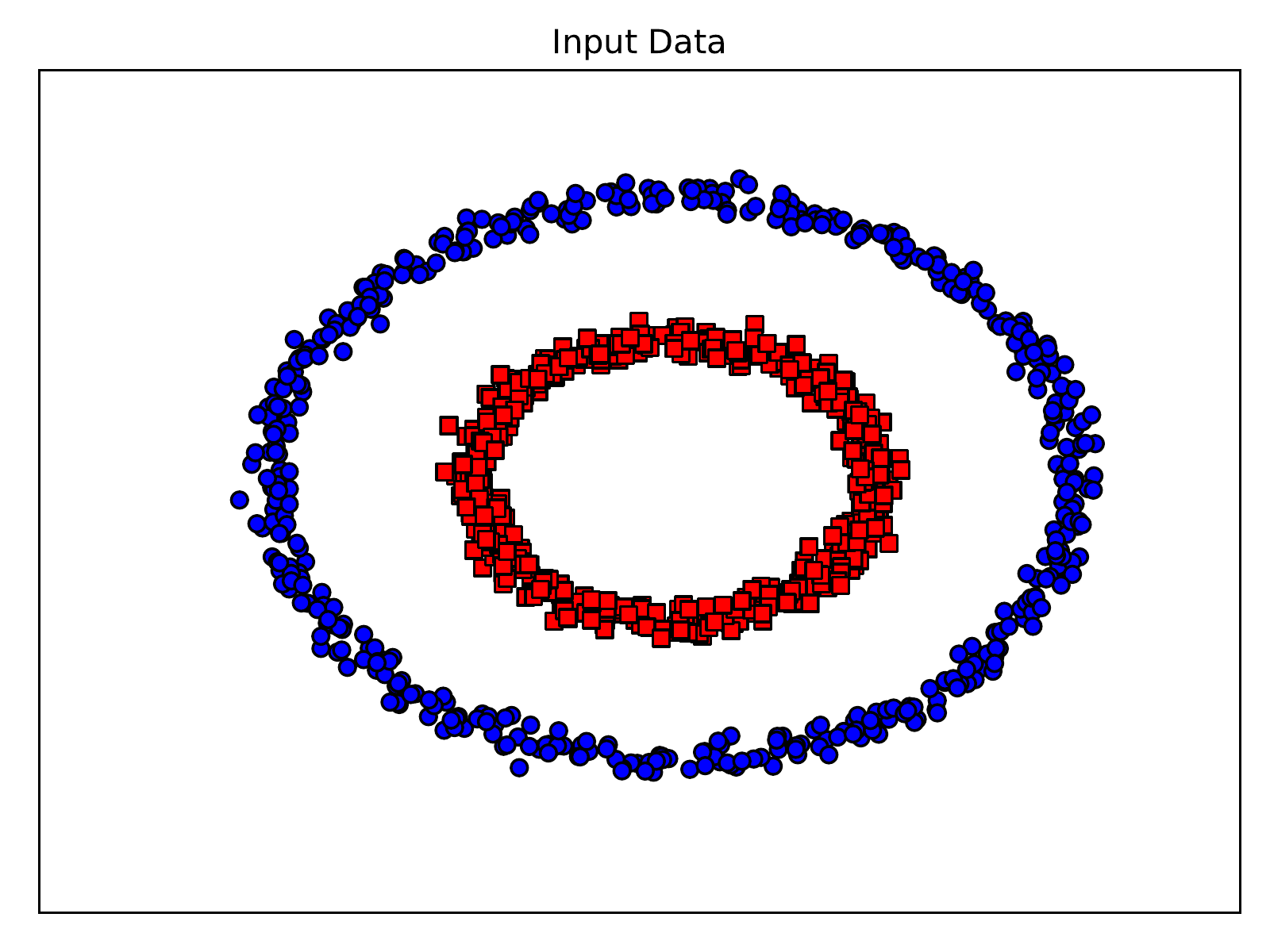}  
%    \caption{Input data}
    \label{fig:sub-first}
  \end{subfigure}
  \begin{subfigure}{.33\textwidth}
    \centering
    % left, lower, right, upper (for trimming)
    \includegraphics[scale=0.3, trim={1cm, 1.5cm, 1cm,
      0.5cm}]{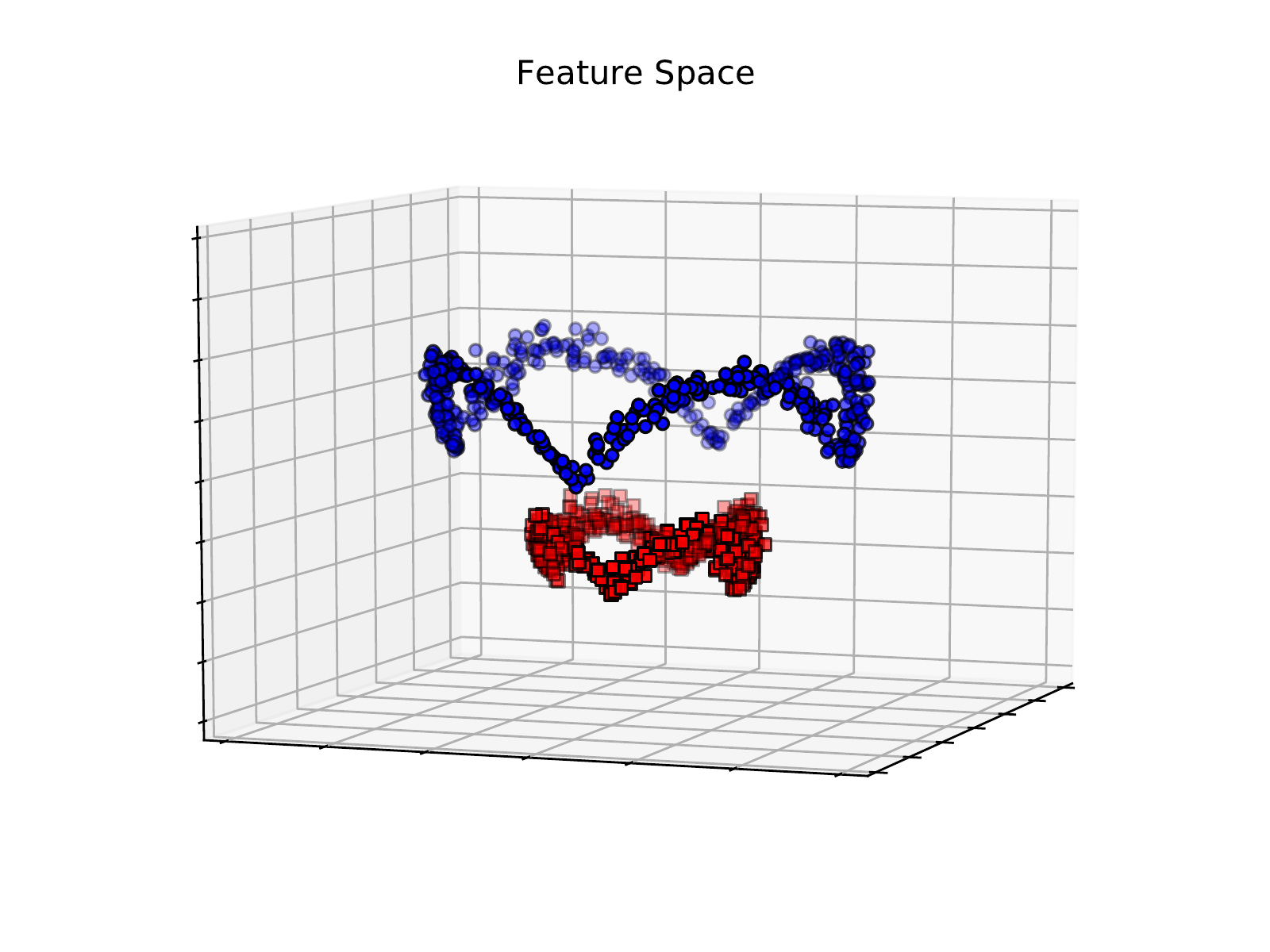}
%    \caption{Feature space}
    \label{fig:sub-second}
  \end{subfigure}
  \begin{subfigure}{.33\textwidth}
    \centering
    \includegraphics[scale=0.25]{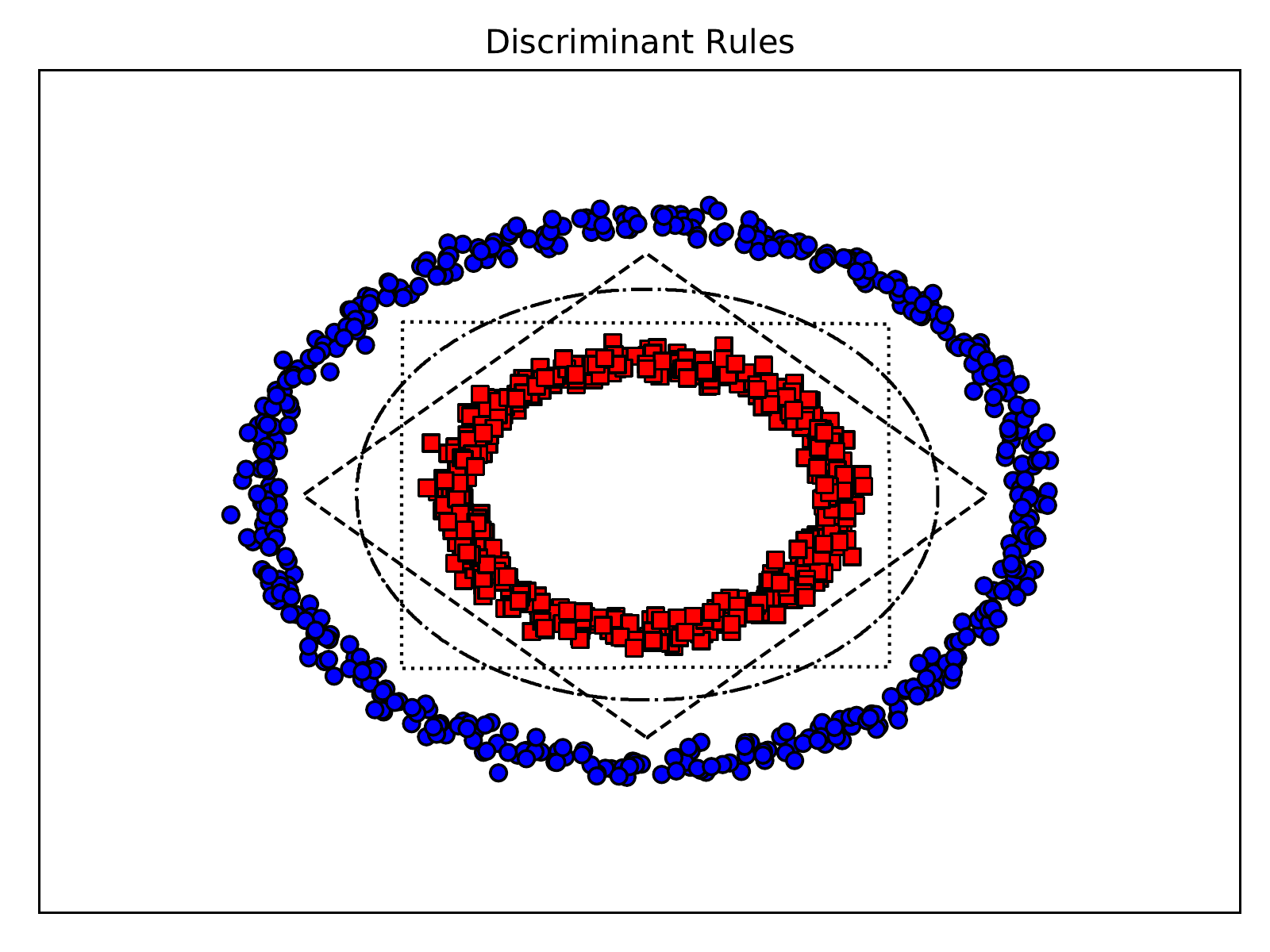}
%    \caption{Discriminant rules}
    \label{fig:sub-second}
  \end{subfigure}
  \\[2mm]
    \begin{subfigure}{.33\textwidth}
    \centering
    \includegraphics[scale=0.25]{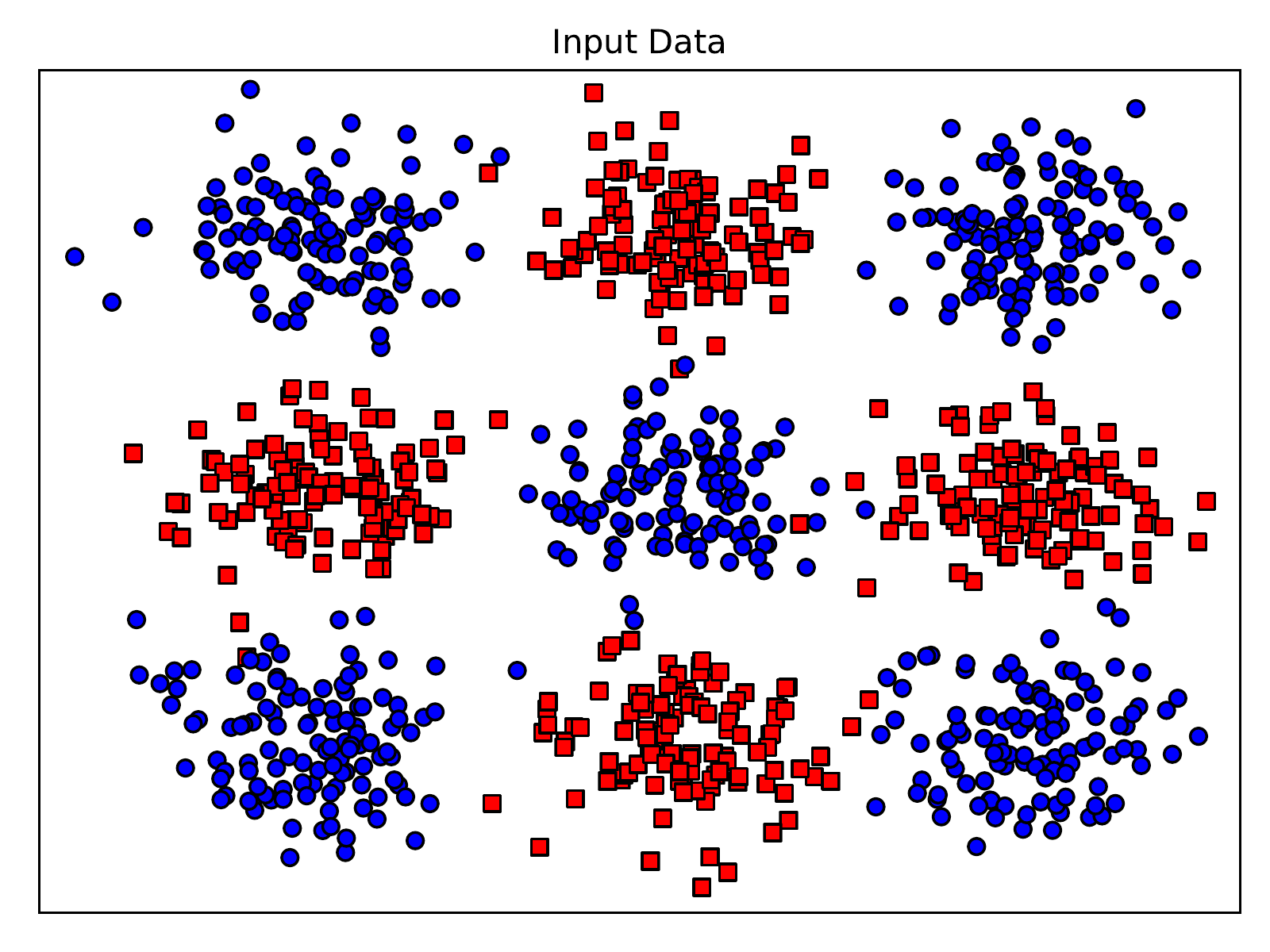}  
%    \caption{Input data}
    \label{fig:sub-first}
  \end{subfigure}
  \begin{subfigure}{.33\textwidth}
    \centering
    % left, lower, right, upper (for trimming)
    \includegraphics[scale=0.3, trim={1cm, 1.5cm, 1cm,
      0.5cm}]{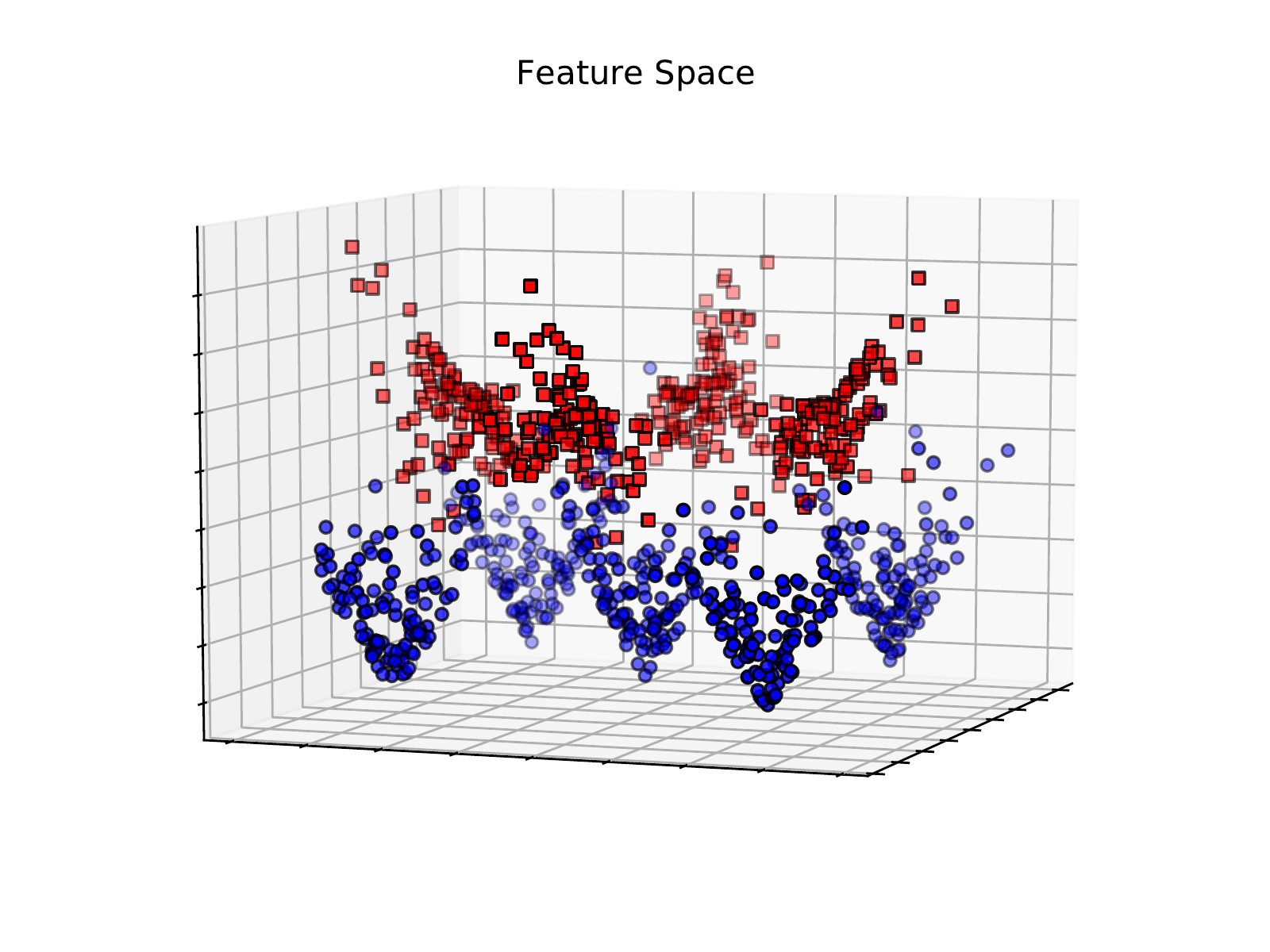}
%    \caption{Feature space}
    \label{fig:sub-second}
  \end{subfigure}
  \begin{subfigure}{.33\textwidth}
    \centering
    \includegraphics[scale=0.25]{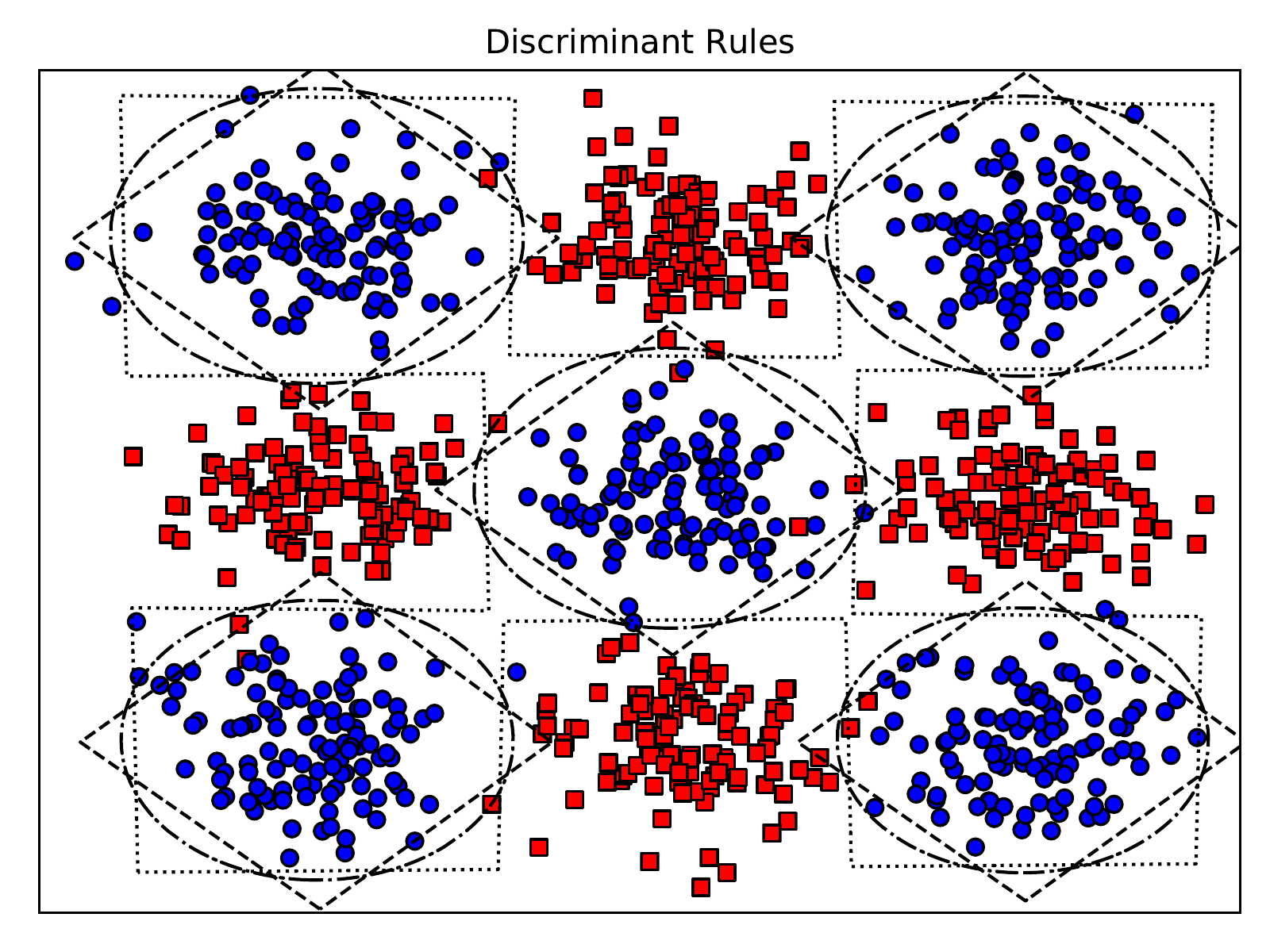}
%    \caption{Discriminant rules}
    \label{fig:sub-second}
  \end{subfigure}
  \caption{Visualization of selecting single (top row) or multiple
    (bottom row) anchor points. The feature spaces are obtained with
    $p=1$. The discriminant rules are given for $p=1$ (diamond), $p=2$
    (ellipsoid) and $p=\infty$ (rectangular). The anchor point in the top row is simply the  average of the whole dataset. The set of anchor points
    in the bottom row are the sample averages within the clusters of
    blue circles.}
  \label{fig:multanchors}
\end{figure}

Up to this point, we have not discussed how to select the anchor point
$\va$. Consider a dataset consisting of the samples $\vx_i \in \RR^d$,
$i \in \CI = \{1,\dots,m\}$. If the user prefers an automatic
selection for the anchor point, then one straightforward choice is the
average of all samples. Depending on the application, the anchor point
could as well be decided by the domain experts. Likewise, the experts
may also propose a set of anchor points $\CA$ instead of just one
point. Given the set of anchor point $\CA$, we may select the closest
point to sample $\vx_i$ as its anchor by evaluating
\begin{equation}
  \va_i = \arg\min_{\va \in \CA} \{\|\vx_i - \va\|_p\}.
\label{eq:mina}
\end{equation}
We can then map $\vx_i$ to a higher-dimensional space with
$\phi_{p,1}(\vx_i~|~\va_i)$ or $\phi_{p,d}(\vx_i~|~\va_i)$. Figure
\ref{fig:multanchors} shows the increase in the dimension as well as
the obtained discriminant rules for two binary classification
problems. The top row is given for a single anchor point, which is
taken as the average of all samples. The bottom row in the same figure
shows the mappings when the set of anchor points $\CA$ is provided by
the user. Here, $\CA$ is constructed with the sample averages of the
clusters of one class (blue circles). Then, the anchor point for each
sample is selected by applying \eqref{eq:mina}. The set $\CA$ can also
be constructed when different samples are known to be associated with
different clusters. These clusters may already shape during data
collection for instance when multiple cohorts, spatial differences,
temporal variations, and so on, are involved. Clearly, different
clusters may as well be formed algorithmically beforehand by using
unsupervised learning methods.

In the main text of our following discussion, we will use a single
anchor point. We have reserved Section \ref{sec:intdim}\footnote{All
  cross references starting with ``S'' refer to the supplementary
  document.}) to introduce different examples, where multiple anchor
points are selected. We will also consider mostly binary
classification problems, since kernels are frequently used within the
well-known Support Vector Machine (SVM) algorithm. The following
proposition formally shows the conditions, under which a linearly
nonseparable binary classification problem can be mapped to a linearly
separable feature space with a single anchor point using the proposed
feature maps.

\begin{prop}
  \label{prop:main}
  Suppose we have a linearly nonseparable dataset with samples
  $\vx_i \in \mathbb{R}^d$ and the corresponding labels
  $y_i \in \{+1,-1\}$ for $i \in \CI$. If we further define two index
  sets $\CI^+ = \{i \in \CI: y_i=+1\}$ and
  $\CI^- = \{i \in \CI : y_i=-1\}$, then using $\phi_{p,1}(\vx~|~\va)$
  returns a linearly separable dataset, if the chosen anchor point
  satisfies
    \begin{equation}
    \label{eq:lem211}
    \min_{i \in \CI^+}\{\|\vx_i-\va\|^p_p\} > \max_{i \in \CI^-} \{\|\vx_i-\va\|^p_p\} ~\text{ or }~
    \max_{i \in \CI^+}\{\|\vx_i-\va\|^p_p\} < \min_{i \in \CI^-} \{\|\vx_i-\va\|^p_p\}.
  \end{equation}
  Likewise, using $\phi_{p,d}(\vx~|~\va)$ returns a linearly separable
  dataset if the chosen anchor point for some dimension
  $\ell \in \{1, \dots, d\}$ satisfies
  \begin{equation}
    \label{eq:lem221}
    \min_{i \in \CI^+}\{|x_{i\ell}-a_\ell|^p\} > \max_{i \in \CI^-} \{| x_{i\ell}-a_\ell|^p\} ~\text{ or }~
    \max_{i \in \CI^+}\{|x_{i\ell}-a_\ell|^p\} < \min_{i \in \CI^-} \{ |x_{i\ell}-a_\ell|^p\}.
  \end{equation}
\end{prop}

The proof of this proposition is given in Section
\ref{sec:proofs}. Although it is a straightforward result, Proposition
\ref{prop:main} provides a clear point of view for the role of the
anchor points. This view actually allows us to conduct a systematic
search for selecting the anchor point. We discuss in
Section \ref{sec:intdim} one such approach yielding complex
discriminant rules particularly useful for multi-class
classification. In any case, we should point out that our numerical
study on benchmark datasets in the next section show that the proposed
feature maps achieve competitive classification accuracies even
without explicitly searching for the \textit{best} anchor point. We
use the sample averages as the anchor points. Even further, we scale
the data so that the sample average becomes the origin. Hence, the
anchor point simply dissapears.

We are ready to present the kernels associated with the proposed
feature maps. Given samples $\vx$ and $\vz$, the proposed kernel
functions are obtained by simply taking the inner product in
higher-dimensional space
\[
  k_{p,1}(\vx,\vz ~|~ \va)= \phi_{p,1}(\vx ~|~ \va)\tr\phi_{p,1}(\vz ~|~ \va) ~~\text{ and }~~
  k_{p,d}(\vx,\vz ~|~ \va)= \phi_{p,d}(\vx ~|~ \va)\tr\phi_{p,d}(\vz ~|~ \va).
\]
Let us now contrast our kernel functions to the well-known radial
basis function and the polynomial kernel function given by
\[
  k_{\gamma}^{\text{RBF}}(\vx, \vz)=e^{-\gamma\|\vx-\vz\|^2}, \gamma>0
  ~~\text{ and }~~ k^{\text{POL}}_q(\vx,\vz)= (\vx\tr\vz + 1)^q, q\in
  \mathbb{N},
\]
respectively. There is no explicit feature map for RBF, since the
input data is mapped to an infinite dimensional space. Although, the
polynomial kernel function is associated with a finite dimensional
space, the resulting dimension can be quite large. Even for
20-dimensional input data, the feature space has more than 200
dimensions for $q=2$. In addition to concerns about interpretability,
these kernels may also slow down the training process. Suppose that we
train a SVM model on the samples $\vx_i \in \mathbb{R}^d$ with the
class labels $y_i\in \{-1, +1\}$ for $i \in \CI$. The training
requires to store an $m \times m$ matrix consisting of kernel function
evaluations of all sample pairs.  Then, a linear system is solved with
this matrix. Thus, the computation time complexity is in the order of
$\CO(m^2)$ to $\CO(m^3)$.  After training, the discriminant rule of
the SVM model becomes
\begin{align*}
  f(\vx) = \sum_{i=1}^{m} \alpha_i y_i k(\vx_i, \vx)
\end{align*}
where $\alpha_i$ values are the weights obtained with training. If the
explicit feature map $\phi_{\bullet}$ is known, then we can also
obtain the weights with
$\vw = \sum_{i=1}^{m} \alpha_i y_i \phi_\bullet(\vx_i)$.

\section{Numerical experiments}
\label{sec:numeric}

In this section, we first compare the performances of the proposed
kernels against linear (LIN), radial basis function (RBF) and
polynmoial (POL) kernels on a set of binary classification datasets
compiled from the literature \cite{Dua:2019,chang2011libsvm}. A
summary of the datasets is given in Section \ref{sec:details}. This
benchmarking study is conducted in terms of prediction accuracy and
training time. We also demonstrate on a particular dataset that our
feature maps constitute a clear choice when compared against two
well-known approximation methods in terms of interpretability.

We have implemented all our kernels in Python\footnote{(GitHub page)
  -- \url{https://github.com/sibirbil/SimpleKernels}} using the
scikit-learn \cite{scikit-learn} package. Since our feature maps are
explicit, we have used the fast linear solver provided with this
package. Thus, we also give the times spent for mappings to give a
fair comparison. For small datasets, we have used 10-fold stratified
cross validation and reported the averages. For the remaining ones, we
have mostly used the test samples provided with the dataset. If a test
set is not available, then we have applied a standard train
(70\%)-test (30\%) split. In all our tests, the anchor point is used
as the sample mean. Thus, scaling the dataset has allowed us to take
$\va = \mathbf{0}$. The other hyperparameters are selected from the
following sets: $\gamma \in \{10^i, i=-5,\dots,4\}$,
$q\in \{2,3,4 \}$, $C \in \{10^i, i=-5,\dots,4\}$. Here $C$ is the SVM
regularization parameter. The best performing parameters are chosen by
applying grid search with stratified two-fold cross validation.

\begin{table}
	\centering	
	\caption{Average accuracies obtained with different methods on small datasets}	
		\label{table:1}
	\begin{tabular}{ l c c c c c c c c c c}
	\toprule
	Datasets			&LIN	&$\phi_{1,1}$	&$\phi_{2,1}$ &$\phi_{1,d}$ &$\phi_{2,d}$	 &POL 	&RBF \\
\midrule                                                                                                                                                                  
Australian		&\textbf{86.81}&	\textbf{86.81}&	86.67&	86.67&	86.09&	84.20&	85.51\\
Fourclass		&76.21&	78.07&	79.35&	77.14&	79.70&	79.36&	\textbf{100.0}\\
Ionosphere		&87.72&	91.44&	92.57&	91.44&	91.73&	91.44&	\textbf{93.72}\\
Heart		 	&84.07&	84.44&	84.07&\textbf{	84.81}&	83.33&	82.96&	84.44\\
Pima  			&\textbf{77.34}&	76.82&	76.68&	76.69&	76.95&	73.43&	76.82\\													
W.Prognostic	&\textbf{81.37}&	79.84&	79.34&	80.37&	77.32&	76.32&	77.74\\
Bupa 			&69.77&	69.5 &	69.18&	\textbf{73.62}&	73.01&	61.85&	72.98 \\
Fertility		&87.00 &	87.00 &	87.00 &	87.00 &	\textbf{88.00} &	86.00 &	86.00\\
W.Diagnostic	&97.07&	96.93&	\textbf{97.22}&	96.78&	96.78&	95.61&	96.93\\

			\bottomrule
		\end{tabular}	

\end{table}

Table \ref{table:1} shows our results on small datasets, where the
acuracy of the best performing method for the corresponding dataset is
written in boldface. Except on Pima and W.Prognostic, the proposed
kernels achieve better average accuracy values than the linear
kernel. For the same two datasets, it is important to note that POL
and RBF also perform worse than the linear kernel. Overall, POL does
not achieve the best accuracy value for any one of the
datasets. Though RBF is the clear winner in two datasets, it is
outperformed by one of our kernels in all other problems. In fact, the
proposed kernels achieve the best predictions on four of the
datasets. Since these are small datasets, the training times of all
kernels are negligible, and hence, we do not report those figures.

\begin{table}	
  \caption{Accuracies obtained with different methods on large
    datasets}
	\label{table:2}
	\centering
	\begin{tabular}{ l c c c c c c c c }
			\toprule
	Datasets				&LIN	&$\phi_{1,1}$	&$\phi_{2,1}$ &$\phi_{1,d}$ &$\phi_{2,d}$	 &POL 	&RBF\\
\midrule                                                
Splice		&	85.29&	86.02&	86.02&	\textbf{91.95}&	89.61&	85.61			&89.93					\\	
Wilt		&   70.60 &	83.60 &	81.20 &	\textbf{85.60} &84.00 &  84.00         &81.80                    	\\
Guide1		&   95.62&	96.25&	96.10 &	96.48&	96.15&     \textbf{96.70}         &96.62           			\\
Spambase	&   92.76&	92.25&	92.90 &	\textbf{94.93}&	93.05& 91.89       		&93.70					\\
Phoneme		&   75.46&	73.61&	74.29&	76.82&	76.57&   78.36           &\textbf{87.55}                      \\
Magic 		&  	79.43&	80.98&	80.30 &	85.61&	84.37&	84.40			&\textbf{87.71}					\\
Adult			&	84.93&	84.93&	84.94&	84.93&	84.92&	84.39	&	\textbf{85.06}						\\
		\bottomrule
	\end{tabular}	
	
\end{table}
      
We report the accuracy values and the training times for large
datasets in Table \eqref{table:2} and Table \eqref{table:3},
respectively. Recall that the training times of the proposed kernels
also include the time spent for explicit mappings. For the large
datasets, we observe that the performance of the linear kernel
degrades. Even if we use only one additional feature with $\phi_{1,1}$
and $\phi_{2,1}$, we obtain better accuracy values than LIN for all
problems but one. For Wilt dataset, this increase goes up to 13\% with
almost no increase in training time. Both POL and RBF also return good
accuracy values than LIN for large datasets. However, Table
\ref{table:3} shows that this improvement comes at a great cost in
training times.  For instance, the accuracy improvement achieved by
POL and RBF on Magic dataset are around $5\%$ and $8\%$,
respectively. The corresponding training times are approximately 60
times and 425 times worse than the linear kernel. All our kernels, on
the other hand, achieve accuracy values on par with POL and RBF in
less than a second.

\begin{table}	
  \caption{Training times in seconds}
	\label{table:3}
	\centering
	\begin{tabular}{ l r r r r r r r r }
		\toprule
		Datasets/Kernels				&LIN	&$\phi_{1,1}$	&$\phi_{2,1}$ &$\phi_{1,d}$ &$\phi_{2,d}$	 &POL 	&RBF\\
		\midrule                                                
	Splice		&	<1 &	<1&	<1		&	<1	&	<1 &	<1 &	<1  \\
	Wilt		&   <1 &   <1 &   <1    &   <1  &   <1 &    <1  &   <1   \\
	Guide1		&   <1 &   <1 &   <1    &   <1  &   <1 &    <1  &   <1   \\
	Spambase	&   <1 &   <1 &   <1    &   <1  &    1.89 &    <1  &   <1   \\
	Phoneme		&   <1 &   <1 &   <1    &   <1  &   <1 &  169.88  &   <1   \\
	Magic 		&  	<1 & <1 &  	<1 &  	<1  &  	<1 &  	425.39 &  	63.94  \\
	Adult		&	<1 &		1.20& 1.21   &  1.39  &	7.43  &	89.20 &	151.36 & \\
		\bottomrule
	\end{tabular}	
	
\end{table}

\begin{figure}
	\begin{subfigure}{.9\textwidth}
		\centering
		\includegraphics[scale=0.3]{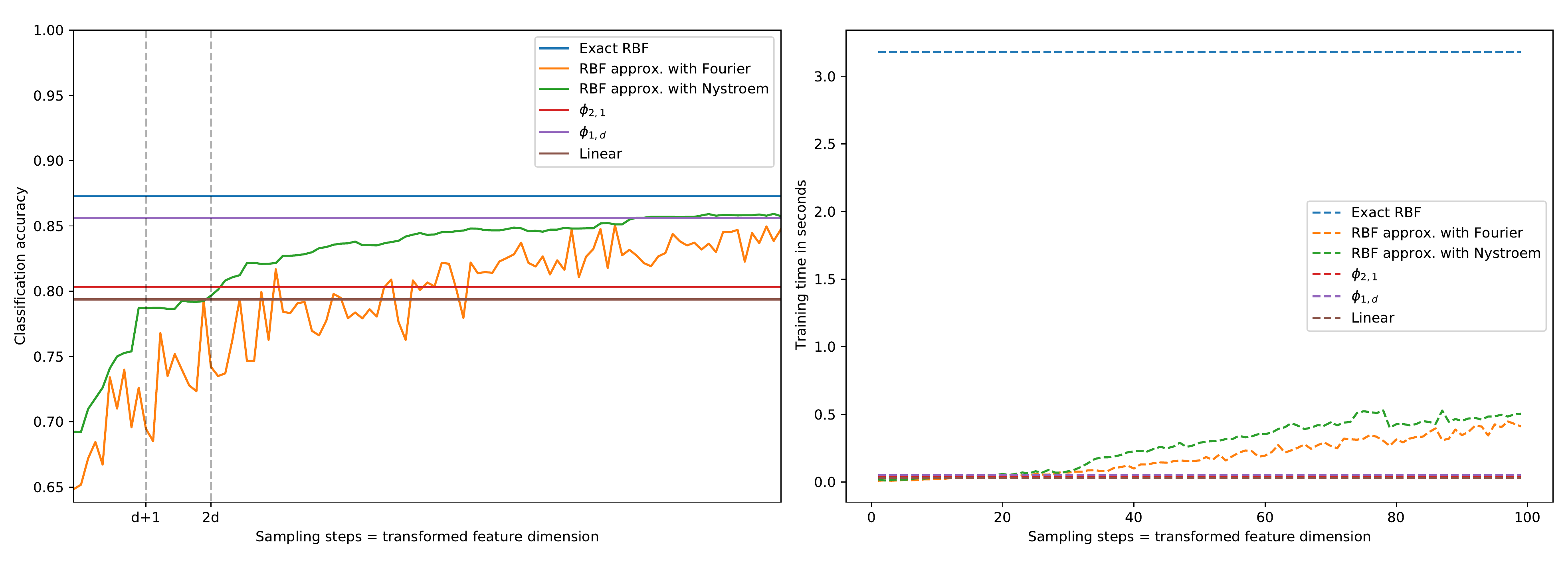}
		%    \caption{Input data}
	\end{subfigure}	
	\caption{Accuracy and training time comparison against
          approximation methods on Magic dataset.}
	\label{fig:approx}
\end{figure}

In the literature, kernel approximation methods are also used to
obtain feature maps with varying dimensions. With these approximation
methods, the accuracy and the computation times oftentimes increase as
the dimension increases. Figure \ref{fig:approx} shows an example
comparison of our kernels against two well-known approximation
approaches, Fourier transform method \cite{rahimi2008random} and
Nyström method \cite{yang2012nystrom}. In these experiments, we have
used the setup for Magic dataset as above with $C=1$ and
$\gamma=0.2$. A practitioner may consider to use the approximation
methods to obtain low (transformed) dimensions for possible
interpretability. However, as the left plot in Figure \ref{fig:approx}
shows, the classification accuracies of both approximation methods are
poor for low transformed feature dimensions. Although an increase in
the dimension increases the accuracy, the training time also goes
up. Our first kernel $\phi_{2,1}$, which adds only one more feature,
performs slightly better than the linear kernel. On the other hand,
when use our second kernel $\phi_{1,d}$ and double the number of
features dimensions, we obtain a classification accuracy value that is
close to RBF. The right plot in Figure \ref{fig:approx} shows that the
training times of our kernels are almost the same as the linear
kernel.

\section{Conclusion}
\label{sec:conclusion}

We propose several low-dimensional kernels that are derived from two
groups of explicit feature maps. The first group expands the feature
space of the samples by adding only one more feature and the second
group doubles the number of features. Both groups append additional
distance-based features to the original list of features. This limited
expansion of the dimension with explicit feature maps brings two
advantages for practitioners: interpretability and fast training.
With our experimental results on several datasets, we have shown that
these advantages do not come at the cost of accuracy. On the contrary,
for several problems, we have obtained better accuracy values than
those of the well-known kernels. Moreover, the training times of our
feature maps have been only a small fraction of the times spent with
those kernels.

The proposed distance-based feature maps are defined with respect to a
single anchor point. In practice, this anchor point can be selected by
the domain experts from a set of candidate points. We also argue that
preprocessing of the samples with clustering methods may also be used
to determine the anchor point. In this work, we have removed the
anchor point from our computational study by scaling the datasets and
taking their sample means as the anchor point. Our numerical results
have shown that even without fine-tuning the selection of the anchor
point, the proposed low-dimensional kernels perform remarkably
well. Nonetheless, we believe that selection of one or more anchor
points can play an important role in prediction accuracy. As we
present in Section \ref{sec:intdim}, our initial experiments show that
the accuracy values can be improved with multiple anchor points
(leading to intermediate dimensions) at almost no cost of additional
training time. Naturally, the resulting feature maps remain
interpretable. The same line of reasoning can also lead to a
systematic search of a set of anchor points for multi-class
classification. These observations shape our future research agenda.

% \section*{Broader impact}

% This work does not present any foreseeable societal consequences.

%\clearpage

%\appendix

\clearpage

\bibliography{mybibfile}

\begin{thebibliography}{19}
\providecommand{\natexlab}[1]{#1}
\providecommand{\url}[1]{\texttt{#1}}
\expandafter\ifx\csname urlstyle\endcsname\relax
  \providecommand{\doi}[1]{doi: #1}\else
  \providecommand{\doi}{doi: \begingroup \urlstyle{rm}\Url}\fi

\bibitem[Avron et~al.(2016)Avron, Sindhwani, Yang, and Mahoney]{avron2016quasi}
H.~Avron, V.~Sindhwani, J.~Yang, and M.~W. Mahoney.
\newblock Quasi-monte carlo feature maps for shift-invariant kernels.
\newblock \emph{The Journal of Machine Learning Research}, 17\penalty0
  (1):\penalty0 4096--4133, 2016.

\bibitem[Cevikalp and Saglamlar(2019)]{cevikalp2019polyhedral}
H.~Cevikalp and H.~Saglamlar.
\newblock Polyhedral conic classifiers for computer vision applications and
  open set recognition.
\newblock \emph{IEEE Transactions on Pattern Analysis and Machine
  Intelligence}, 2019.

\bibitem[Cevikalp and Triggs(2017)]{Cevikalp_2017_CVPR}
H.~Cevikalp and B.~Triggs.
\newblock Polyhedral conic classifiers for visual object detection and
  classification.
\newblock In \emph{The IEEE Conference on Computer Vision and Pattern
  Recognition (CVPR)}, July 2017.

\bibitem[Chang and Lin(2011)]{chang2011libsvm}
C.-C. Chang and C.-J. Lin.
\newblock Libsvm: A library for support vector machines.
\newblock \emph{ACM Transactions on Intelligent Systems and Technology (TIST)},
  2\penalty0 (3):\penalty0 1--27, 2011.

\bibitem[Cimen and Ozturk(2019)]{cimenOpcf}
E.~Cimen and G.~Ozturk.
\newblock O-pcf algorithm for one-class classification.
\newblock \emph{Optimization Methods and Software}, pages 1--15, 2019.

\bibitem[Cimen et~al.(2018)Cimen, Ozturk, and Gerek]{cimen2018incremental}
E.~Cimen, G.~Ozturk, and O.~N. Gerek.
\newblock Incremental conic functions algorithm for large scale classification
  problems.
\newblock \emph{Digital Signal Processing}, 77:\penalty0 187--194, 2018.

\bibitem[Dua and Graff(2017)]{Dua:2019}
D.~Dua and C.~Graff.
\newblock {UCI} machine learning repository, 2017.
\newblock URL \url{http://archive.ics.uci.edu/ml}.

\bibitem[Gasimov and Ozturk(2006)]{gasimov2006separation}
R.~N. Gasimov and G.~Ozturk.
\newblock Separation via polyhedral conic functions.
\newblock \emph{Optimization Methods and Software}, 21\penalty0 (4):\penalty0
  527--540, 2006.

\bibitem[Hamid et~al.(2014)Hamid, Xiao, Gittens, and DeCoste]{hamid2014compact}
R.~Hamid, Y.~Xiao, A.~Gittens, and D.~DeCoste.
\newblock Compact random feature maps.
\newblock In \emph{International Conference on Machine Learning}, pages 19--27,
  2014.

\bibitem[Huang et~al.(2013)Huang, Mehrkanoon, and Suykens]{huang2013support}
X.~Huang, S.~Mehrkanoon, and J.~A. Suykens.
\newblock Support vector machines with piecewise linear feature mapping.
\newblock \emph{Neurocomputing}, 117:\penalty0 118 -- 127, 2013.
\newblock ISSN 0925-2312.

\bibitem[Kar and Karnick(2012)]{kar2012random}
P.~Kar and H.~Karnick.
\newblock Random feature maps for dot product kernels.
\newblock In \emph{Artificial Intelligence and Statistics}, pages 583--591,
  2012.

\bibitem[Maji and Berg(2009)]{maji2009max}
S.~Maji and A.~C. Berg.
\newblock Max-margin additive classifiers for detection.
\newblock In \emph{2009 IEEE 12th International Conference on Computer Vision},
  pages 40--47. IEEE, 2009.

\bibitem[Ozturk and Ciftci(2015)]{ozturk2015clustering}
G.~Ozturk and M.~T. Ciftci.
\newblock Clustering based polyhedral conic functions algorithm in
  classification.
\newblock \emph{Journal of Industrial \& Management Optimization}, 11\penalty0
  (3):\penalty0 921, 2015.

\bibitem[Pedregosa et~al.(2011)Pedregosa, Varoquaux, Gramfort, Michel, Thirion,
  Grisel, Blondel, Prettenhofer, Weiss, Dubourg, Vanderplas, Passos,
  Cournapeau, Brucher, Perrot, and Duchesnay]{scikit-learn}
F.~Pedregosa, G.~Varoquaux, A.~Gramfort, V.~Michel, B.~Thirion, O.~Grisel,
  M.~Blondel, P.~Prettenhofer, R.~Weiss, V.~Dubourg, J.~Vanderplas, A.~Passos,
  D.~Cournapeau, M.~Brucher, M.~Perrot, and E.~Duchesnay.
\newblock Scikit-learn: Machine learning in {P}ython.
\newblock \emph{Journal of Machine Learning Research}, 12:\penalty0 2825--2830,
  2011.

\bibitem[Pham and Pagh(2013)]{pham2013fast}
N.~Pham and R.~Pagh.
\newblock Fast and scalable polynomial kernels via explicit feature maps.
\newblock In \emph{Proceedings of the 19th ACM SIGKDD International Conference
  on Knowledge Discovery and Data mining}, pages 239--247, 2013.

\bibitem[Rahimi and Recht(2008)]{rahimi2008random}
A.~Rahimi and B.~Recht.
\newblock Random features for large-scale kernel machines.
\newblock In \emph{Advances in Neural Information Processing Systems}, pages
  1177--1184, 2008.

\bibitem[Si et~al.(2017)Si, Hsieh, and Dhillon]{si2017memory}
S.~Si, C.-J. Hsieh, and I.~S. Dhillon.
\newblock Memory efficient kernel approximation.
\newblock \emph{The Journal of Machine Learning Research}, 18\penalty0
  (1):\penalty0 682--713, 2017.

\bibitem[Vempati et~al.(2010)Vempati, Vedaldi, Zisserman, and
  Jawahar]{vempati2010generalized}
S.~Vempati, A.~Vedaldi, A.~Zisserman, and C.~Jawahar.
\newblock Generalized {RBF} feature maps for efficient detection.
\newblock In \emph{BMVC}, pages 1--11, 2010.

\bibitem[Yang et~al.(2012)Yang, Li, Mahdavi, Jin, and Zhou]{yang2012nystrom}
T.~Yang, Y.-F. Li, M.~Mahdavi, R.~Jin, and Z.-H. Zhou.
\newblock Nystr{\"o}m method vs random fourier features: A theoretical and
  empirical comparison.
\newblock In \emph{Advances in Neural Information Processing Systems}, pages
  476--484, 2012.

\end{thebibliography}

\clearpage

\renewcommand\thesection{S.\arabic{section}}
\setcounter{section}{0}
\renewcommand\thetable{S.\arabic{table}}
\setcounter{table}{0}
% \renewcommand{\thealgocf}{SM.\arabic{algocf}}
% \setcounter{algocf}{0}
% \pagenumbering{arabic}
% \setcounter{page}{1}
% \resetlinenumber

\begin{center}
  {\Large Supplementary Material for \\[2mm]
    ``Low-dimensional Interpretable Kernels with \\ Conic Discriminant
    Functions for Classification''}
\end{center}

\bigskip

Note that most of the cross references in this supplementary file
refer to the original manuscript. Therefore, clicking on those
references may not take you to the desired location.  All the results
and the figures both in the main document and in this supplementary
document can be reproduced with the scripts that are provided in the
accompanying compressed file.

\section{Omitted proof}
\label{sec:proofs}

\proofnum{of Proposition \ref{prop:main}: }{Suppose that the first
  part of \eqref{eq:lem211} holds and define 
  \[
    \mu = \min_{i \in \CI^+}\{\|\vx_i-\va\|^p_p\} ~\text{ and }~ \nu =
    \max_{i \in \CI^-} \{\|\vx_i-\va\|^p_p\}.
  \]
 If we take $\bar{\vw}_{1:d+1} = (\mathbf{0}\tr_{1:d}, 1)\tr$, then
  for any $j \in \CI^+$ and $k \in \CI^-$ we have
  \[
    \bar{\vw}\tr_{1:d+1}\phi_{p,1}(\vx_j~|~\va) = \|\vx_j-\va\|^p_p \geq
    \mu > \nu \geq \bar{\vw}\tr_{1:d+1}\phi_{p,1}(\vx_k~|~\va) = \|\vx_k-\va\|^p_p.
  \]
  We have just obtained the strict linear separation. The same line of
  arguments can be followed to show the rest of the proposition after
  selecting the particular dimension $\ell \in \{1, \dots,
  d\}$. Suppose that the first part of \eqref{eq:lem221} holds and
  define
  \[
    \mu_\ell = \min_{i \in \CI^+}\{ |x_{i\ell}-a_\ell|^p\} ~\text{
      and }~ \nu_\ell = \max_{i \in \CI^-} \{|x_{i\ell}-a_\ell|^p\}.
  \]
  Take $\bar{\vw}_{1:2d} = (\mathbf{0}\tr_{1:d}, \ve_\ell\tr)\tr$, where
  $\ve_\ell$ shows the $\ell$th unit vector. Then for any
  $j \in \CI^+$ and $k \in \CI^-$ we have
  \[
    \bar{\vw}\tr_{1:2d}\phi_{p,d}(\vx_j~|~\va) = |x_{j\ell}-a_\ell|^p\geq
    \mu_\ell > \nu_\ell \geq \bar{\vw}\tr_{1:2d}\phi_{p,d}(\vx_k~|~\va) = |x_{k\ell}-a_\ell|^p.
  \]
  We have just obtained the strict linear separation. The second parts
  of both \eqref{eq:lem211} and \eqref{eq:lem221} can simply be
  obtained by reversing the roles of $\CI^+$ and $\CI^-$.  

}

\section{Intermediate dimensions and multi-class case}
\label{sec:intdim}

Proposition \ref{prop:main} provides the conditions for achieving a
linear separation by using the proposed feature maps. However,
satisfying these conditions with a single anchor point may be
difficult. Consider instead the case when a set of anchor points is
provided. The following result shows the conditions under which we can
obtain a linearly separable dataset. Here, each sample is associated
with one of the anchor points in the set. The proof of this result can
be shown with the same arguments as in the proof of Proposition
\ref{prop:main}.

\begin{corollary}
  \label{cor:1}
  In addition to the setup as in Proposition \ref{prop:main}, suppose
  that we also have a set of anchor points $\CA$. If we further define
  for all $i \in \CI$, the feature maps $\phi_{p,1}(\vx_i~|~\va_i)$
  with $\va_i = \arg\min_{\va \in \CA} \{\|\vx_i - \va\|_p\}$, then we
  obtain a linearly separable dataset when the chosen anchor points
  satisfy
  \begin{equation}
	\label{eq:cor1}
	\min_{i \in \CI^+}\{\|\vx_i-\va_i\|^p_p\} > \max_{i \in \CI^-} \{\|\vx_i-\va_i\|^p_p\} ~\text{ or }~
	\max_{i \in \CI^+}\{\|\vx_i-\va_i\|^p_p\} < \min_{i \in \CI^-} \{\|\vx_i-\va_i\|^p_p\}.
	\end{equation}
\end{corollary}

\begin{figure}[h]
		\centering
	\begin{subfigure}{.40\textwidth}
		\centering
		\includegraphics[scale=0.3]{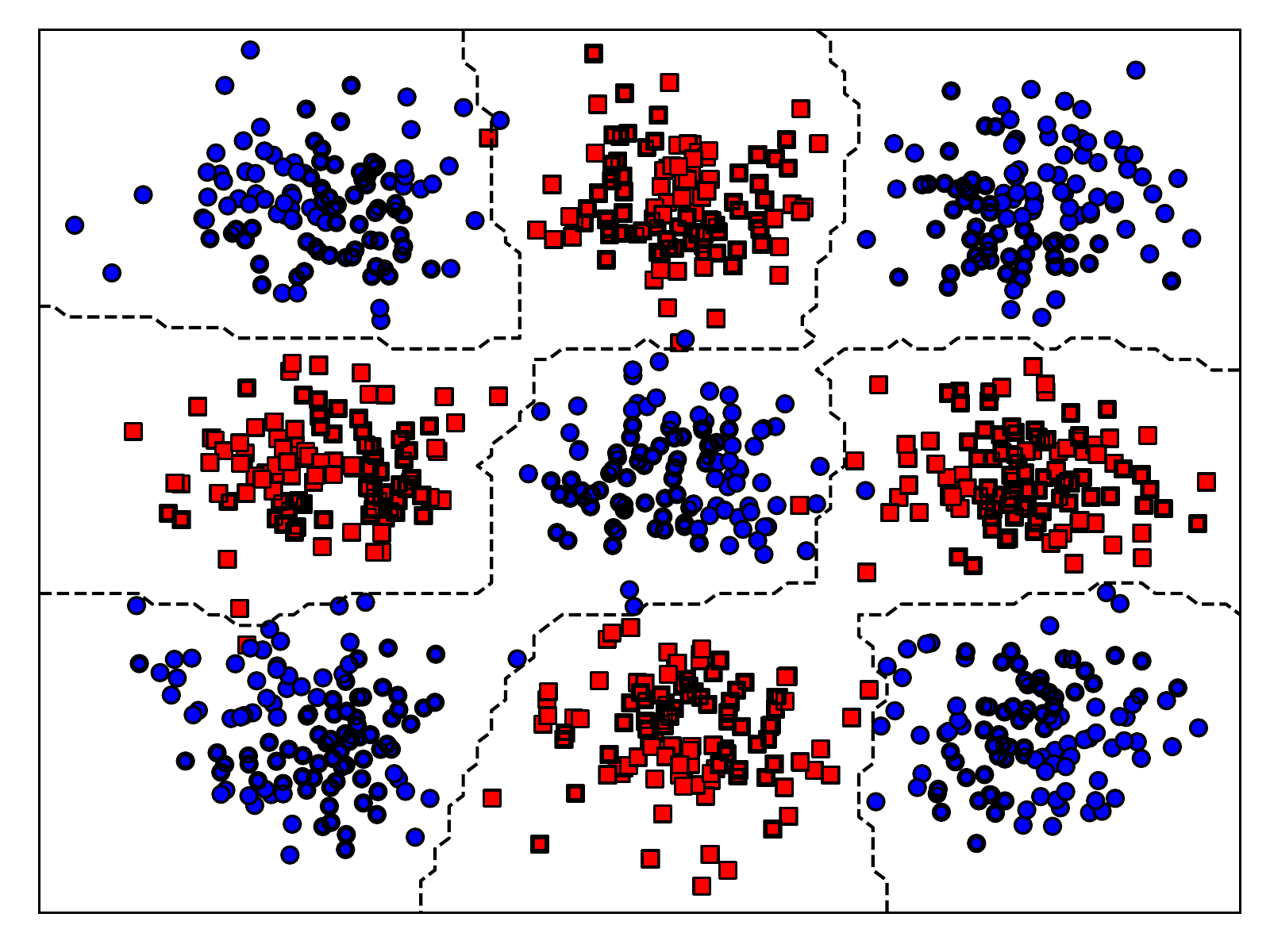}
		%    \caption{Input data}
	\end{subfigure}
	\begin{subfigure}{.40\textwidth}
		\centering
	\includegraphics[scale=0.3]{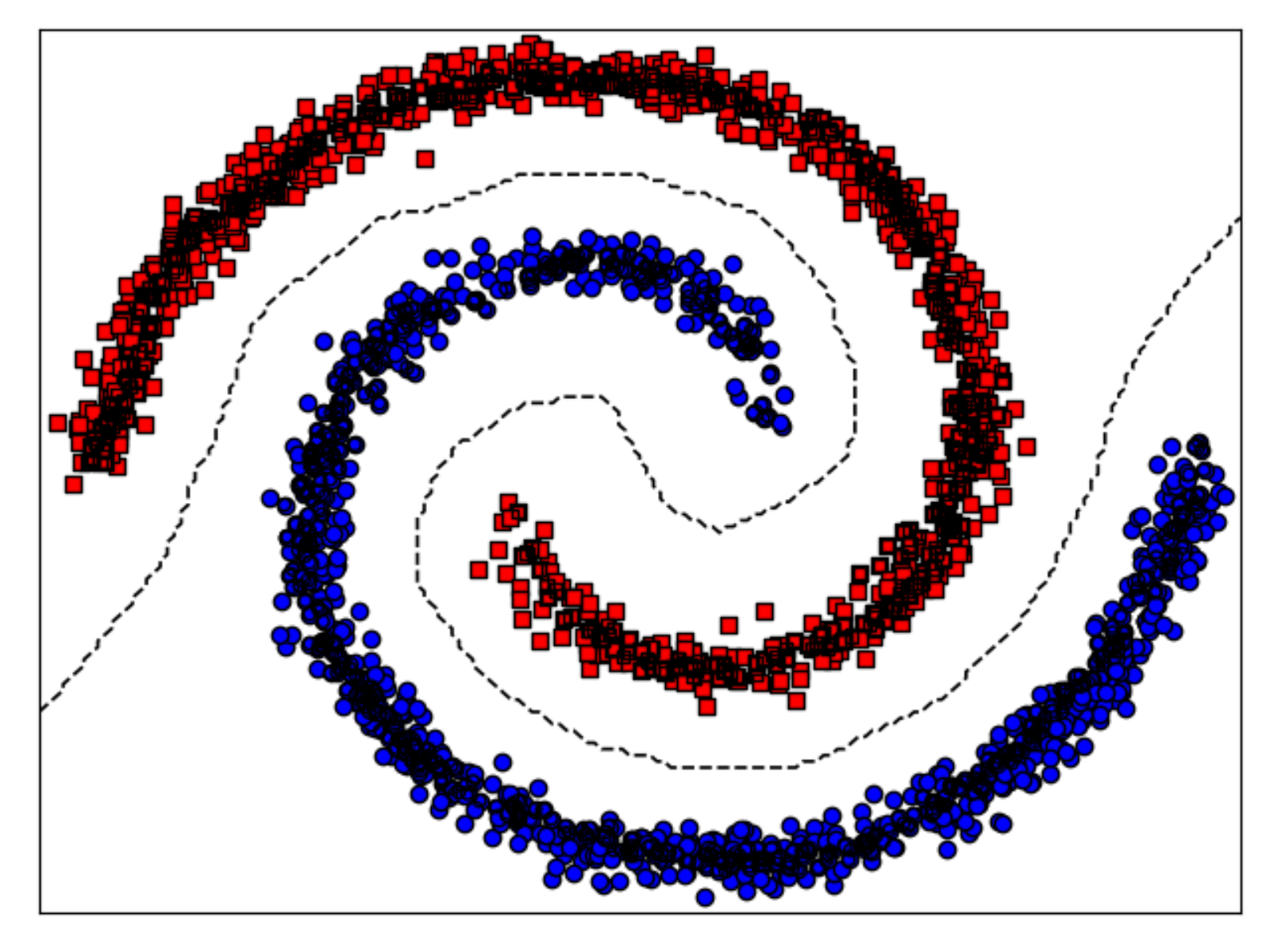}
	%    \caption{Input data}
\end{subfigure}
\caption{The discriminant functions obtained with selecting two sets
  of anchor points from different classes.}
 \label{fig:complex}
\end{figure}

Using a set of anchor points brings flexibility. For instance, one may
simply select $\CA$ as all the samples from one of the classes. This
selection immediately satisfies the conditions stated in Corollary
\ref{cor:1}. However, it is important to keep in mind that such a set
is likely to lead to overfitting. Therefore, we next introduce another
approach, where two sets of anchor points $\CA_1$ and $\CA_2$ are
formed with selection of samples from both classes. Then, the feature
map that increases the dimension by two for $i \in \CI$ simply becomes
\[
  \phi_{p,2}(\vx_i~|~\va^{(1)}_i,\va_i^{(2)}) = (\vx_i, \| \vx_i - \va^{(1)}_i\|^p_p,\|\vx_i -
  \va^{(2)}_i\|^p_p)\tr,
\]
where $\va^{(1)}_i = \arg\min_{\va \in \CA_1} \{\|\vx_i - \va\|_p\}$
and $\va^{(2)}_i = \arg\min_{\va \in \CA_2} \{\|\vx_i -
\va\|_p\}$. Figure \ref{fig:complex} gives two illustrations of
discriminant functions that are obtained with this feature map.

\begin{figure}[h]
	\centering
	\begin{subfigure}{.4\textwidth}
		\centering
		\includegraphics[scale=0.35]{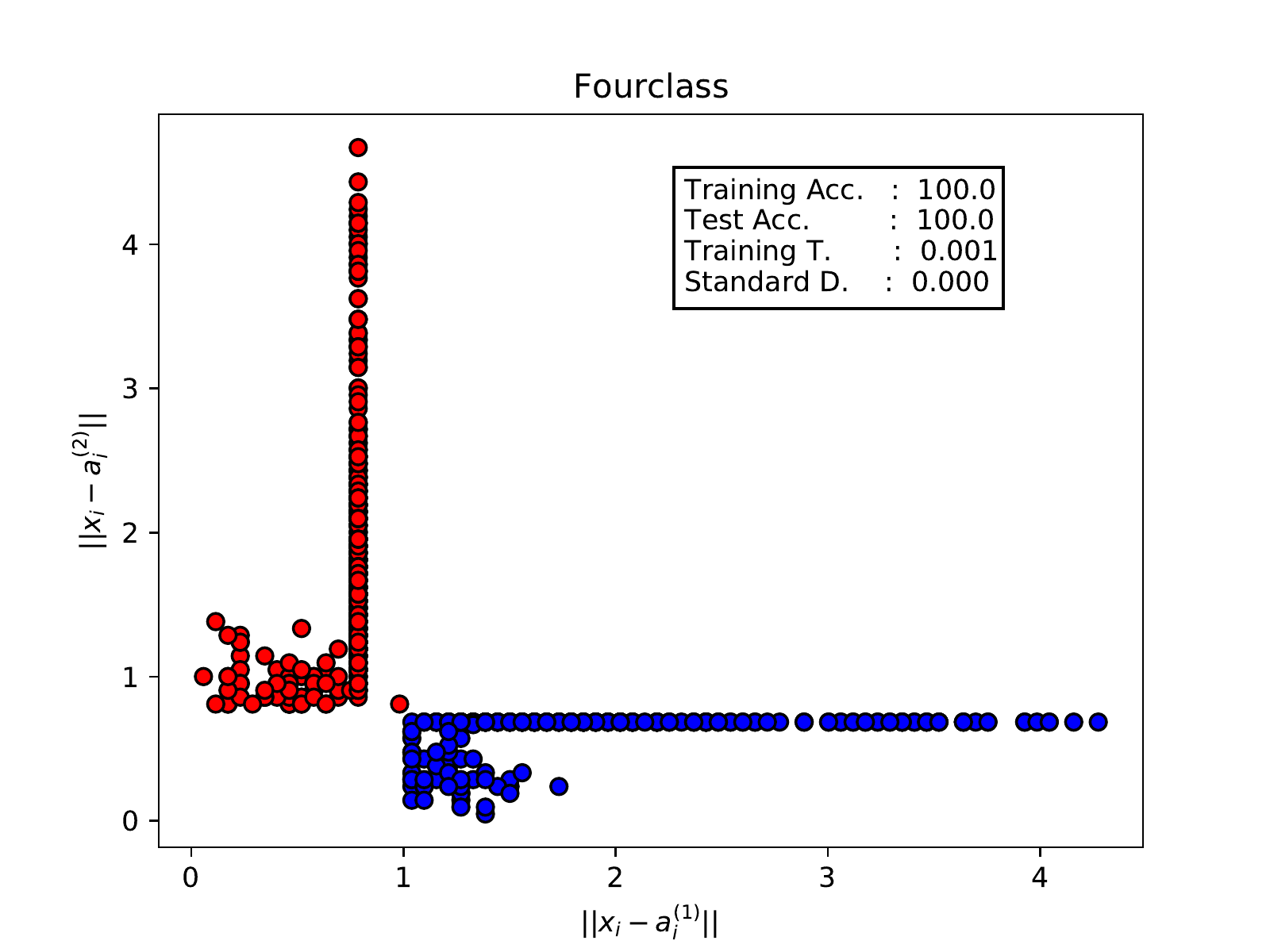}  
		%    \caption{Input data}
	\end{subfigure}
	\begin{subfigure}{.4\textwidth}
		\centering
		% left, lower, right, upper (for trimming)
		\includegraphics[scale=0.35]{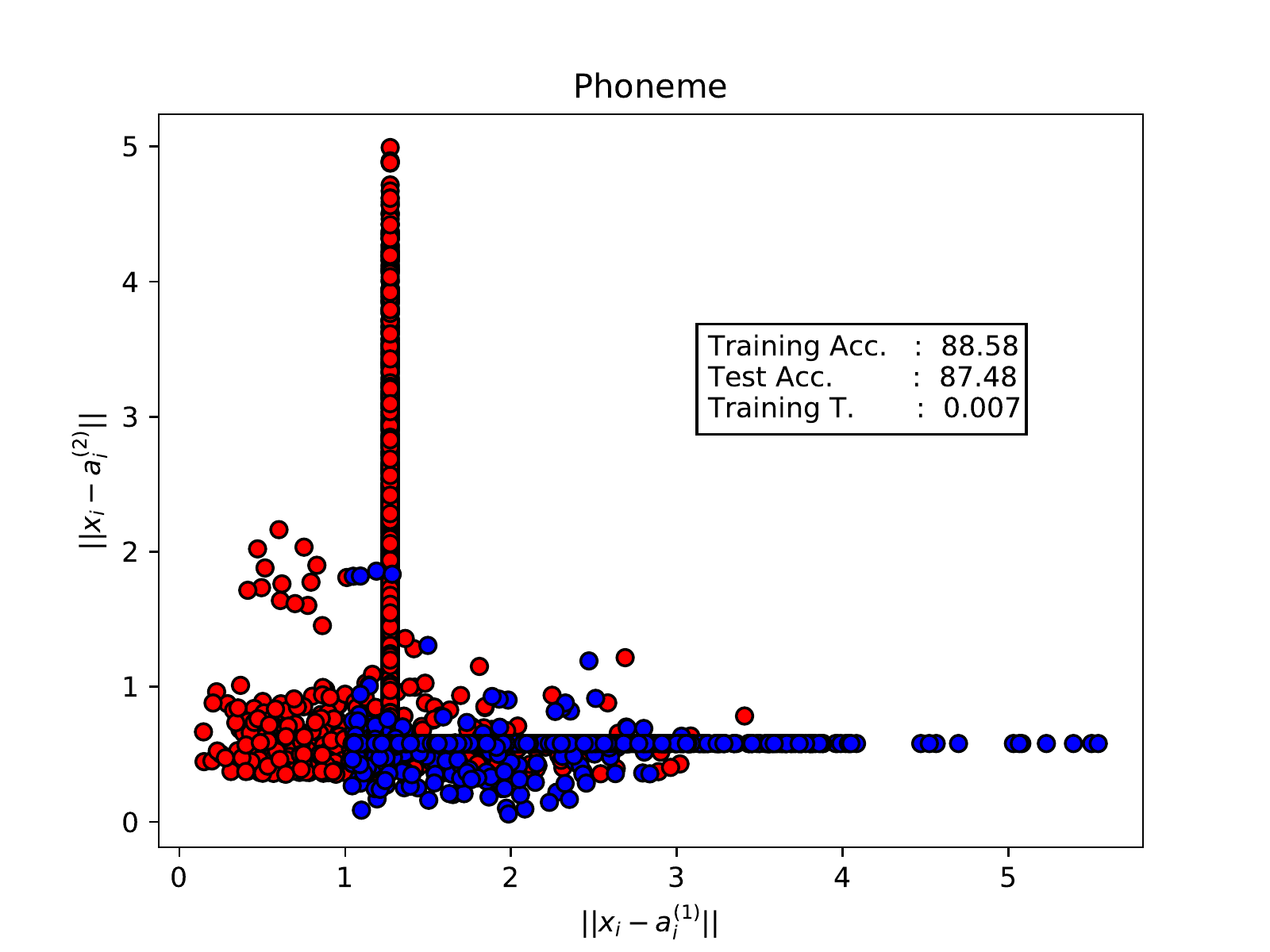}
		%    \caption{Feature space}
	\end{subfigure}
	\caption{The last two components of $\phi_{1,2}$ and the
          performance metrics. In both tests, we have used the same
          setup as in in our numerical experiments section. The best
          classification accuracy results that we have obtained before
          on Fourclass and Phonome datasets were $79.70\%$ and
          $78.36\%$, respectively.}
	\label{fig:multidplus2}
\end{figure}

Then comes the important issue of forming the anchor sets. In our
numerical experiments, we have observed that selecting the closest or
the farthest sample as the anchor point of a particular sample does
not perform well as they are likely to be the support vectors or the
outliers (a possible increase in model variance). Thus, one can select
the samples within a certain distance that is bounded from above and
below. Figure \ref{fig:multidplus2} illustrates the distributions of
distances that can be used with the new feature map
$\phi_{p,2}$. Among these values, we have used only a lower bound to
remove the outliers shown in the south-west corners of both
plots. Then, the remaining samples are used to form the anchor point
sets. We have applied a simple grid search to decide the lower
bound. As the reported metrics in each plot show, we have improved our
previous accuracies significantly with this simple approach.

Note that while searching for the anchor points, the training time
remains the same with each trial. This is, however, not the case with
other maps in the literature, since they all increase the dimension of
the feature space. Moreover, the search for anchor points could also
be invaluable for the practitioners, since they can gain important
insights about the \textit{critical} samples and the structure of
their datasets.

\begin{figure}[h]
	\centering
	\begin{subfigure}{.9\textwidth}
		\centering
		\includegraphics[scale=0.4]{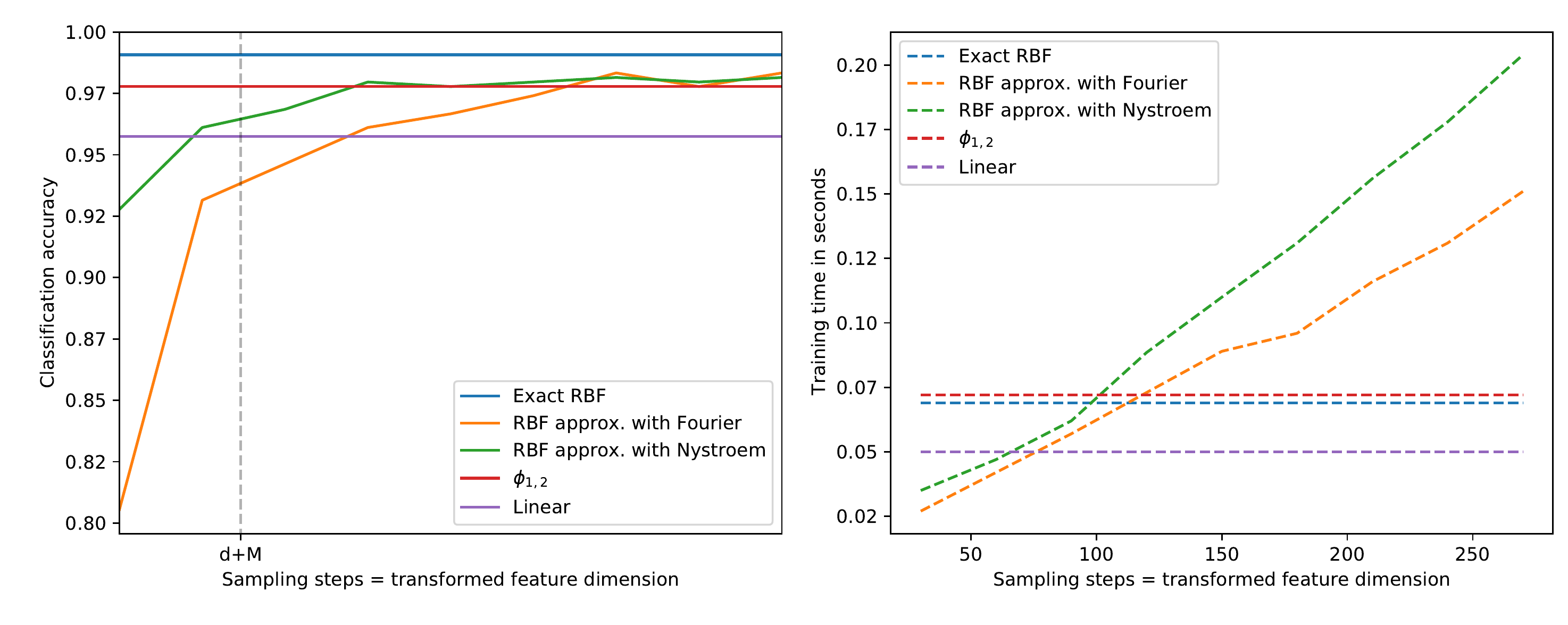}  
		%    \caption{Input data}
	\end{subfigure}
	\caption{Accuracy and training time comparison against
          approximation methods on Pen Digits dataset ($M=10$). This
          dataset comes with scikit-learn \cite{scikit-learn}.}
	\label{fig:approxmulti}
\end{figure}

Finally, we give a discussion on multi-class classification. Consider
a dataset with $M$ classes, \textit{i.e.},
$y_i \in \{1, 2, \dots, M\}$, $i \in \CI$. We next introduce for
$i \in \CI$ the feature map
\[
  \phi_{p,M}(\vx_i~|~\va_i^{(1)}, \dots, \va_i^{(M)}) = (\vx_i, \| \vx_i -
  \va_i^{(1)}\|^p_p, \dots, \|\vx_i - \va_i^{(M)}\|^p_p)\tr,
\]
where $\CA_1, \dots, \CA_M$ are the anchor sets and
$\va^{(j)}_i = \arg\min_{\va \in \CA_j} \{\|\vx_i - \va\|_p\}$ for
$j=1, \dots, M$. When there are two classes, we clearly obtain
$\phi_{p,2}$. Figure \ref{fig:approxmulti} shows a comparison of
$\phi_{p,M}$ against two well-known kernel approximation methods,
Fourier transform \cite{rahimi2008random} and Nyström
\cite{yang2012nystrom}. The sets of anchor points are selected with a
similar framework that described for the binary problems. As we have
observed in Section \ref{sec:numeric}, the proposed multi-class
feature map performs better than the approximation methods and gives a
classification accuracy on par with RBF.

\section{Relation to polyhedral conic functions}
\label{sec: pcf}

Our main discriminant functions have ties with so-called conic
functions in the literature. The very first work on polyhedral conic
functions (PCFs) is given by \citet{gasimov2006separation}. This work
is followed by a series of studies proposing various PCF-based
classifiers \cite{ozturk2015clustering, Cevikalp_2017_CVPR,
  cimen2018incremental, cimenOpcf}. In all these works, PCFs are
considered as constraints within optimization problems solved for
classification. In fact, our current work establishes that the
classifiers based on PCFs can be simply considered as kernel methods
using explicit feature maps.

% In this section, first we give brief
% summaries of the researches and then, make simple observations that
% show these classifiers are kernel methods that use explicit feature
% maps. In the end of this section, we also discuss differences between
% our study and existing polyhedral conic classifiers.

\citet{gasimov2006separation} have proposed the function
$g:\mathbb{R}^d\to\mathbb{R}$ given by
\begin{equation}
\label{eqn:pcf}
g(\vx)=\vw\tr(\vx-\va)+w_{d+1}\|\vx-\va\|_1-b,
\end{equation}
where $\vw,\va \in \mathbb{R}^d$ and $w_{d+1}, b \in \mathbb{R}$. Note
that except the displacement with $\mathbf{a}$ in the first term, this
function is equivalent to one of our discriminant functions,
$f_{1,1}$. The function in \eqref{eqn:pcf} is polyhedral conic, since
its graph is a cone with vertex at
$(\va, -b)\in \mathbb{R}^d\times\mathbb{R}$ and all its sublevel sets
are polyhedrons \cite[Lemma 2.1]{gasimov2006separation}.  The authors
propose an iterative classification algorithm, which is based on
solving a series of linear programming problems. At each iteration, a
new vector $\va$ is selected randomly, and the corresponding
constraint using function \eqref{eqn:pcf} is used to form a new linear
programming model. Due to this iterative structure and random
selection, the training time of the proposed classifier is not
comparable with the state-of-the-art classifiers, and its prediction
performance depends on the choice of $\va$ vectors. Later, the authors
have also tried clustering methods to select these vectors
\cite{ozturk2015clustering} and considered different $\ell_p$-norms
\cite{cimen2018incremental}. It is important to note that the term
\emph{polyhedral conic separation} defined in
\cite{gasimov2006separation} corresponds to a linear separation in the
$d+1$ dimensional feature space. This is actually the point of view
that we advocate in Section \ref{sec:proposed}.

As an extension of \cite{gasimov2006separation},
\citet{Cevikalp_2017_CVPR} consider function
$h:\mathbb{R}^d\to\mathbb{R}$ given by
\begin{equation}
\label{eqn:epcf}
h(\vx)=\vw\tr(\vx-\va)+ \sum_{\ell=1}^d w_{d+\ell}|x_\ell-a_\ell|-b,
\end{equation}
where $\vw,\va \in \mathbb{R}^d$ and $b, w_{d+\ell} \in \mathbb{R}$
for $\ell=1, \dots, d$.  This time, our discriminant function,
$f_{1,d}$ is equivalent to \eqref{eqn:epcf} except the displacement
with $\mathbf{a}$ in the first term. Unlike others,
\citet{Cevikalp_2017_CVPR} use both \eqref{eqn:pcf} and
\eqref{eqn:epcf} in SVM optimization model and report results on a set
of visual object detection problems. In a follow-up study
\cite{cevikalp2019polyhedral}, the authors have also taken the
squares of the absolute value terms in \eqref{eqn:epcf} and obtained
ellipsoidal conic functions. They have also considered \eqref{eqn:pcf}
with $\ell_2$-norm instead of $\ell_1$-norm. In the same study, the
authors have also hinted that the samples are ``explicitly mapped to a
higher feature space.''  However, they have not followed with this
line of thought to discuss a general framework for the feature maps
and the associated kernels as we do here.

As a last note, we point out that none of the work above has
considered the intermediate dimensions, nor given a formal discussion
on conditions for linear separation (see Proposition \ref{prop:main}
and Corollary \ref{cor:1}), which allows us to consider a systematic
way to explore multiple anchor points.

\section{Details of the datasets}
\label{sec:details}

In this section, we give list the properties of the datasets that we
have used in our numerical experiments. All these datasets are also
provided with the accompanying compressed file.

\begin{table}[h]
	\centering
	\caption{Properties of the datasets}
	\label{tab:datasets}
	\begin{tabular}{ l r r r}
	\toprule
	Datasets & $m$ & $d$ & \\
	\midrule
	Australian 		&690 & 14\\
	Fourclass 		&862 & 2\\
	Ionosphere  	&351 & 34\\
	Heart			&270 & 13\\
	Pima Indians Diabetes	&768 &8\\
	Wisconsin Prognosis			&194& 33\\
	Bupa Liver			&341 &6\\
	Fertility		&100 & 9\\
	Wisconsin Diagnosis			&683 & 10\\
	Splice			& 3175&60\\
	Wilt			&4889 &5\\
	SVM Guide1		& 7089&4\\
	Spambase		&4601 & 57\\	
	Phoneme			&5404 & 5\\
	Magic 			&19020 &10\\
          Adult &48852&123 \\
          \bottomrule
\end{tabular}
\end{table}
 \end{document}